# Probabilistic Theorem Proving

**Vibhav Gogate** and **Pedro Domingos**
University of Washington
Computer Science & Engineering
Seattle, WA 98195, USA
{vgogate,pedrod}@cs.washington.edu

## Abstract

Many representation schemes combining first-order logic and probability have been proposed in recent years. Progress in unifying logical and probabilistic inference has been slower. Existing methods are mainly variants of lifted variable elimination and belief propagation, neither of which take logical structure into account. We propose the first method that has the full power of both graphical model inference and first-order theorem proving (in finite domains with Herbrand interpretations). We first define probabilistic theorem proving, their generalization, as the problem of computing the probability of a logical formula given the probabilities or weights of a set of formulas. We then show how this can be reduced to the problem of lifted weighted model counting, and develop an efficient algorithm for the latter. We prove the correctness of this algorithm, investigate its properties, and show how it generalizes previous approaches. Experiments show that it greatly outperforms lifted variable elimination when logical structure is present. Finally, we propose an algorithm for approximate probabilistic theorem proving, and show that it can greatly outperform lifted belief propagation.

## 1 INTRODUCTION

Unifying first-order logic and probability enables uncertain reasoning over domains with complex relational structure, and is a long-standing goal of AI. Proposals go back to at least Nilsson [27], with substantial progress within the UAI community starting in the 1990s (e.g., [1, 19, 40]), and added impetus from the new field of statistical relational learning starting in the 2000s [16]. Many well-developed representations now exist (e.g., [9, 14, 23]), but the state of inference is less advanced. For the most part, inference is still carried out by converting models to propositional form (e.g., Bayesian networks) and then applying standard propositional algorithms. This typically incurs an exponential blowup in the time and space cost of inference, and forgoes one of the chief attractions of first-order logic: the ability to perform *lifted* inference, i.e., reason over large domains in time independent of the number of objects they contain, using techniques like resolution theorem proving [32].

In recent years, progress in lifted probabilistic inference has picked up. An algorithm for lifted variable elimination was proposed by Poole [29] and extended by de Salvo Braz [10] and others. Lifted belief propagation was introduced by Singla and Domingos [38] and extended by others (e.g., [21, 36]). These algorithms often yield impressive efficiency gains compared to propositionalization, but still fall well short of the capabilities of first-order theorem proving, because they ignore logical structure, treating potentials as black boxes. This paper proposes the first full-blown probabilistic theorem prover, capable of exploiting both lifting and logical structure, and having standard theorem proving and standard graphical model inference as special cases.

Our solution is obtained by reducing probabilistic theorem proving (PTP) to lifted weighted model counting. We first do the corresponding reduction for the propositional case, extending previous work by Darwiche [6] and Sang et al. [34] (see also [4]). We then lift this approach to the first-order level, and refine it in several ways. We show that our algorithm can be exponentially more efficient than first-order variable elimination, and is never less efficient (up to constants). For domains where exact inference is not feasible, we propose a sampling-based approximate version of our algorithm. Finally, we report experiments in which PTP greatly outperforms first-order variable elimination and belief propagation, and discuss future research directions.

## 2 LOGIC AND THEOREM PROVING

We begin with a brief review of propositional logic, first-order logic and theorem proving [15]. The simplest formulas in propositional logic are *atoms*: individual symbols representing propositions that may be true of false in a given world. More complex formulas are recursively built up from atoms and the logical connectives ¬ (negation), ∧ (conjunction), ∨ (disjunction), ⇒ (implication) and ⇔

**Algorithm 1** TP(KB $K$, query $Q$)
--------
$K_Q \leftarrow K \cup \{\neg Q\}$
**return** $\neg$SAT(CNF($K_Q$))
--------

(equivalence). For example, $\neg$A $\vee$ (B $\wedge$ C) is true iff A is false or B and C are true. A *knowledge base (KB)* is a set of logical formulas. The fundamental problem in logic is determining *entailment*, and algorithms that do this are called *theorem provers*. A knowledge base $K$ entails a query formula $Q$ iff $Q$ is true in all worlds where all the formulas in $K$ are true, a *world* being an assignment of truth values to all atoms. A world is a *model* of a KB iff the KB is true in it. Theorem provers typically first convert $K$ and $Q$ to *conjunctive normal form (CNF)*. A CNF formula is a conjunction of *clauses*, each of which is a disjunction of *literals*, each of which is an atom or its negation. For example, the CNF of $\neg$A $\vee$ (B $\wedge$ C) is ($\neg$A $\vee$ B) $\wedge$ ($\neg$A $\vee$ C). A *unit clause* consists of a single literal. Entailment can then be computed by adding $\neg Q$ to $K$ and determining whether the resulting KB $K_Q$ is *satisfiable*, i.e., whether there exists a world where all clauses in $K_Q$ are true. If not, $K_Q$ is unsatisfiable, and $K$ entails $Q$. Algorithm 1 shows this basic theorem proving schema. CNF($K$) converts $K$ to CNF, and SAT($C$) returns True if $C$ is satisfiable and False otherwise.

The earliest theorem prover is the Davis-Putnam algorithm (henceforth called DP) [8]. It makes use of the *resolution* rule: if a KB contains the clauses $A_1 \vee \ldots \vee A_n$ and $B_1 \vee \ldots \vee B_m$, where the $a$'s and $b$'s represent literals, and some literal $A_i$ is the negation of some literal $B_j$, then the clause $A_1 \vee \ldots \vee A_{i-1} \vee A_{i+1} \vee \ldots \vee A_n \vee B_1 \vee \ldots \vee B_{j-1} \vee B_{j+1} \vee \ldots \vee B_m$ can be added to the KB. For each atom A in the KB, DP resolves every pair of clauses $C_1, C_2$ in KB such that $C_1$ contains A and $C_2$ contains $\neg$A, adds the result to the KB, and deletes $C_1$ and $C_2$. If at some point the empty clause is derived, the KB is unsatisfiable, and the query formula (previously negated and added to the KB) is therefore proven. As Dechter [11] points out, DP is in fact just the variable elimination algorithm for the special case of 0-1 potentials.

Modern propositional theorem provers use the DPLL algorithm [7], a variant of DP that replaces the elimination step with a *splitting* step: instead of eliminating all clauses containing the chosen atom A, resolve all clauses in the KB with A, simplify and recurse, and do the same with $\neg$A. If both recursions fail, the KB is unsatisfiable. DPLL has linear space complexity, compared to exponential for Davis-Putnam. DPLL is the basis of the algorithms in this paper.

First-order logic inherits all the features of propositional logic, and in addition allows atoms to have internal structure. An atom is now a predicate symbol, representing a relation in the domain of interest, followed by a parenthesized list of variables and/or constants, representing objects. For example, Friends(Anna,x) is an atom. A *ground atom* has only constants as arguments. First-order logic has two additional connectives, $\forall$ (universal quantification) and $\exists$ (existential quantification). For example, $\forall$x Friends(Anna,x) means that Anna is friends with everyone, and $\exists$x Friends(Anna,x) means that Anna has at least one friend. In this paper, we assume that domains are finite (and therefore function-free) and that there is a one-to-one mapping between constants and objects in the domain (Herbrand interpretations).

As long as the domain is finite, first-order theorem proving can be carried out by *propositionalization*: creating atoms from all possible combinations of predicates and constants, and applying a propositional theorem prover. However, this is potentially very inefficient. A more sophisticated alternative is first-order resolution [32], which proceeds by resolving pairs of clauses and adding the result to the KB until the empty clause is derived. Two first-order clauses can be resolved if they contain complementary literals that *unify*, i.e., there is a *substitution* of variables by constants or other variables that makes the two literals identical up to the negation sign. Conversion to CNF is carried out as before, with the additional step of removing all existential quantifiers by a process called *skolemization*.

First-order logic allows knowledge to be expressed vastly more concisely than propositional logic. For example, the rules of chess can be stated in a few pages in first-order logic, but require hundreds of thousands in propositional logic. Probabilistic logical languages extend this power to uncertain domains. The goal of this paper is to similarly extend the power of first-order theorem proving.

## 3 PROBLEM DEFINITION

Following Nilsson [27], we define probabilistic theorem proving as the problem of determining the probability of an arbitrary query formula $Q$ given a set of logical formulas $F_i$ and their probabilities $P(F_i)$. For the problem to be well defined, the probabilities must be consistent, and Nilsson [27] provides a method for verifying consistency. Probabilities estimated by maximum likelihood from an observed world are guaranteed to be consistent [13]. In general, a set of formula probabilities does not specify a complete joint distribution over the atoms appearing in them, but one can be obtained by making the *maximum entropy* assumption: the distribution contains no information beyond that specified by the formula probabilities [27]. Finding the maximum entropy distribution given a set of formula probabilities is equivalent to learning a maximum-likelihood log-linear model whose features are the formulas; many algorithms for this purpose are available (iterative scaling, gradient descent, etc.) [13].

We call a set of formulas and their probabilities together with the maximum entropy assumption a *probabilistic knowledge base (PKB)*. Equivalently, a PKB can be directly defined as a log-linear model with the formulas as features and the corresponding weights or potential values. Potentials are the most convenient form, since they allow determinism (0-1 probabilities) without recourse to infinity. If **x**

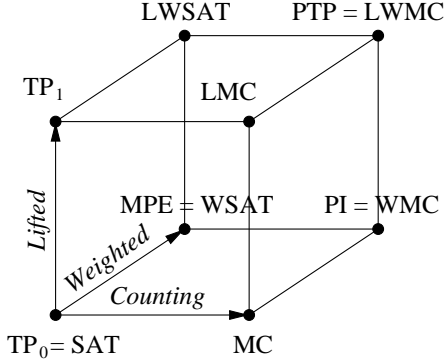

Figure 1: Inference problems addressed in this paper. $TP_o$ and $TP_1$ is propositional and first-order theorem proving respectively, PI is probabilistic inference (computing marginals), MPE is computing the most probable explanation, SAT is satisfiability, MC is model counting, W is weighted and L is lifted. A = B means A can be reduced to B.

is a world and $\Phi_i(\mathbf{x})$ is the potential corresponding to formula $F_i$, by convention (and without loss of generality) we let $\Phi_i(\mathbf{x}) = 1$ if $F_i$ is true, and $\Phi_i(\mathbf{x}) = \phi_i \geq 0$ if the formula is false. *Hard* formulas have $\phi_i = 0$ and *soft* formulas have $\phi_i > 0$. In order to compactly subsume standard probabilistic models, we interpret a universally quantified formula as a set of features, one for each grounding of the formula, as in Markov logic [14]. A PKB $\{(F_i, \phi_i)\}$ thus represents the joint distribution

$$P(\mathbf{X}=\mathbf{x}) = \frac{1}{Z} \prod_i \phi_i^{n_i(\mathbf{x})}, \quad (1)$$

where $n_i(\mathbf{x})$ is the number of false groundings of $F_i$ in $\mathbf{x}$, and $Z$ is a normalization constant (the *partition function*). We can now define PTP succinctly as follows:

**Probabilistic theorem proving (PTP)**
*Input:* Probabilistic KB $K$ and query formula $Q$
*Output:* $P(Q|K)$

If all formulas are hard, a PKB reduces to a standard logical KB. Determining whether a KB $K$ logically entails a query $Q$ is equivalent to determining whether $P(Q|K) = 1$ [14]. Graphical models are easily converted into equivalent PKBs [4]. Conditioning on evidence is done by adding the corresponding hard ground atoms to the PKB, and the conditional marginal of an atom is computed by issuing the atom as the query. Thus PTP has both logical theorem proving and inference in graphical models as special cases. In this paper we solve PTP by reducing it to lifted weighted model counting. *Model counting* is the problem of determining the number of worlds that satisfy a KB. *Weighted* model counting can be defined as follows [4]. Assign a weight to each literal, and let the weight of a world be the product of the weights of the literals that are true in it. Then weighted model counting is the problem of determining the sum of the weights of the worlds that satisfy a KB:

**Weighted model counting (WMC)**
*Input:* CNF $C$ and set of literal weights $W$
*Output:* Sum of weights of worlds that satisfy $C$

Figure 1 depicts graphically the set of inference problems

---

**Algorithm 2** WCNF(PKB $K$)
**for all** $(F_i, \phi_i) \in K$ s.t. $\phi_i > 0$ **do**
  $K \leftarrow K \cup \{(F_i \Leftrightarrow A_i, 0)\} \setminus \{(F_i, \phi_i)\}$
$C \leftarrow \text{CNF}(K)$
**for all** $\neg A_i$ literals **do** $W_{\neg A_i} \leftarrow \phi_i$
**for all** other literals $L$ **do** $W_L \leftarrow 1$
**return** $(C, W)$

**Algorithm 3** PTP(PKB $K$, query $Q$)
$K_Q \leftarrow K \cup \{(Q, 0)\}$
**return** WMC(WCNF($K_Q$))/WMC(WCNF($K$))

---

addressed by this paper. Generality increases in the direction of the arrows. We first propose an algorithm for propositional weighted model counting and then lift it to first-order. The resulting algorithm is applicable to all the problems in the diagram. (Weighted SAT/MPE inference requires replacing sums with maxes and an additional trace-back step, but we do not pursue this here; cf. Park [28], and de Salvo Braz [10] on the lifted case.)

## 4 PROPOSITIONAL CASE

This section generalizes the Bayesian network inference techniques in Darwiche [5] and Sang et al. [34] to arbitrary propositional PKBs, evidence, and query formulas. Although this is of value in its own right, its main purpose is to lay the groundwork for the first-order case.

The probability of a formula is the sum of the probabilities of the worlds that satisfy it. Thus to compute the probability of a formula $Q$ given a PKB $K$ it suffices to compute the partition function of $K$ with and without $Q$ added as a hard formula:

$$P(Q|K) = \frac{\sum_{\mathbf{x}} 1_Q(\mathbf{x}) \prod_i \Phi_i(\mathbf{x})}{Z(K)} = \frac{Z(K \cup \{(Q,0)\})}{Z(K)} \quad (2)$$

where $1_Q(\mathbf{x})$ is the indicator function (1 if $Q$ is true in $\mathbf{x}$ and 0 otherwise).

In turn, the computation of partition functions can be reduced to weighted model counting using the procedure in Algorithm 2. This replaces each soft formula $F_i$ in $K$ by a corresponding hard formula $F_i \Leftrightarrow A_i$, where $A_i$ is a new atom, and assigns to every $\neg A_i$ literal a weight of $\phi_i$ (the value of the potential $\Phi_i$ when $F_i$ is false).

**Theorem 1** $Z(K) = \text{WMC}(\text{WCNF}(K))$.

*Proof.* If a world violates any of the hard clauses in $K$ or any of the $F_i \Leftrightarrow A_i$ equivalences, it does not satisfy $C$ and is therefore not counted. The weight of each remaining world $\mathbf{x}$ is the product of the weights of the literals that are true in $\mathbf{x}$. By the $F_i \Leftrightarrow A_i$ equivalences and the weights assigned by WCNF($K$), this is $\prod_i \Phi_i(\mathbf{x})$. (Recall that $\Phi_i(\mathbf{x}) = 1$ if $F_i$ is true in $\mathbf{x}$ and $\Phi_i(\mathbf{x}) = \phi_i$ otherwise.) Thus $\mathbf{x}$'s weight is the unnormalized probability of $\mathbf{x}$ under $K$. Summing these yields the partition function $Z(K)$.  □

Equation 2 and Theorem 1 lead to Algorithm 3 for PTP.

**Algorithm 4** WMC(CNF $C$, weights $W$)
// *Base case*
**if** all clauses in $C$ are satisfied **then**
　**return** $\prod_{A \in \mathbf{A}(C)} (W_A + W_{\neg A})$
**if** $C$ has an empty unsatisfied clause **then return** 0
// *Decomposition step*
**if** $C$ can be partitioned into CNFs $C_1, \ldots, C_k$ sharing no
　atoms **then**
　**return** $\prod_{i=1}^{k} \text{WMC}(C_i, W)$
// *Splitting step*
Choose an atom $A$
**return** $W_A \text{WMC}(C|A; W) + W_{\neg A} \text{WMC}(C|\neg A; W)$

(Compare with Algorithm 1.) WMC($C, W$) can be any weighted model counting algorithm [4]. Most model counters are variations of Relsat, itself an extension of DPLL [3]. Relsat splits on atoms until the CNF is decomposed into sub-CNFs sharing no atoms, and recurses on each sub-CNF. This decomposition is crucial to the efficiency of the algorithm. In this paper we use a weighted version of Relsat, shown in Algorithm 4. $\mathbf{A}(C)$ is the set of atoms that appear in $C$. $C|A$ denotes the CNF obtained by resolving each clause in $C$ with $A$, which results in removing $\neg A$ from all clauses it appears in, and setting to *Satisfied* all clauses in which $A$ is true. Notice that, unlike in DPLL, satisfied clauses cannot simply be deleted, because we need to keep track of which atoms they are over for model counting purposes. However, they can be ignored in the decomposition step, since they introduce no dependencies. The atom to split on in the splitting step can be chosen using various heuristics [35].

**Theorem 2** *Algorithm WMC($C,W$) correctly computes the weighted model count of CNF $C$ under literal weights $W$.*

*Proof sketch.* If all clauses in $C$ are satisfied, all assignments to the atoms in $C$ satisfy it, and the corresponding total weight is $\prod_{A \in \mathbf{A}(C)} (W_A + W_{\neg A})$. If $C$ has an empty unsatisfied clause, it is unsatisfiable given the truth assignment so far, and the corresponding weighted count is 0. If two CNFs share no atoms, the WMC of their conjunction is the product of the WMCs of the individual CNFs. Splitting on an atom produces two disjoint sets of worlds, and the total WMC is therefore the sum of the WMCs of the two sets, weighted by the corresponding literal's weight. □

## 5 FIRST-ORDER CASE

We now lift PTP to the first-order level. We consider first the case of PKBs without existential quantifiers. Algorithms 2 and 3 remain essentially unchanged, except that formulas, literals and CNF conversion are now first-order. In particular, for Theorem 1 to remain true, each new atom $A_i$ in Algorithm 2 must now consist of a new predicate symbol followed by a parenthesized list of the variables and constants in the corresponding formula $F_i$. The proof of the first-order version of the theorem then follows by propositionalization. Lifting Algorithm 4 is the focus of the rest of this section.

**Algorithm 5** LWMC(CNF $C$, substs. $S$, weights $W$)
// *Lifted base case*
**if** all clauses in $C$ are satisfied **then**
　**return** $\prod_{A \in \mathbf{A}(C)} (W_A + W_{\neg A})^{n_A(S)}$
**if** $C$ has an empty unsatisfied clause **then return** 0
// *Lifted decomposition step*
**if** there exists a lifted decomposition $\{C_{1,1}, \ldots, C_{1,m_1},$
　$\ldots, C_{k,1}, \ldots, C_{k,m_k}\}$ of $C$ under $S$ **then**
　**return** $\prod_{i=1}^{k} [\text{LWMC}(C_{i,1}, S, W)]^{m_i}$
// *Lifted splitting step*
Choose an atom $A$
Let $\{\Sigma_{A,S}^{(1)}, \ldots, \Sigma_{A,S}^{(l)}\}$ be a lifted split of $A$ for $C$ under $S$
**return** $\sum_{i=1}^{l} n_i W_A^{t_i} W_{\neg A}^{f_i} \text{LWMC}(C|\sigma_j; S_j, W)$
　where $n_i$, $t_i$, $f_i$, $\sigma_j$ and $S_j$ are as in Proposition 3

We begin with some necessary definitions. A *substitution constraint* is an expression of the form $\text{x} = \text{y}$ or $\text{x} \neq \text{y}$, where x is a variable and y is either a variable or a constant. (Much richer substitution constraint languages are possible, but we adopt the simplest one that allows PTP to subsume both standard function-free theorem proving and first-order variable elimination.) Two literals are *unifiable* under a set of substitution constraints $S$ if there exists at least one ground literal consistent with $S$ that is an instance of both, up to the negation sign. A $(C, S)$ pair, where $C$ is a first-order CNF whose variables have been standardized apart and $S$ is a set of substitution constraints, represents the ground CNF obtained by replacing each clause in $C$ with the conjunction of its groundings that are consistent with the constraints in $S$. For example, using upper case for constants and lower case for variables, if $C = \text{R}(\text{B},\text{C}) \wedge (\neg \text{R}(\text{x},\text{y}) \vee \text{S}(\text{y},\text{z}))$ and $S = \{\text{x}=\text{y}, \text{z} \neq \text{B}\}$, $(C, S)$ represents the ground CNF $\text{R}(\text{B},\text{C}) \wedge (\neg \text{R}(\text{B},\text{B}) \vee \text{S}(\text{B},\text{C})) \wedge (\neg \text{R}(\text{C},\text{C}) \vee \text{S}(\text{C},\text{C}))$. Clauses with equality substitution constraints can be abbreviated in the obvious way (e.g., $\text{T}(\text{x},\text{y},\text{z})$ with $\text{x}=\text{y}$ and $\text{z}=\text{C}$ can be abbreviated as $\text{T}(\text{x},\text{x},\text{C})$).

We lift the base case, decomposition step, and splitting step of Algorithm 4 in turn. The result is shown in Algorithm 5. In addition to the first-order CNF $C$ and weights on first-order literals $W$, LWMC takes as an argument an initially empty set of substitution constraints $S$ which, similar to logical theorem proving, is extended along each branch of the inference as the algorithm progresses.

### 5.1 LIFTING THE BASE CASE

The base case changes only by raising each first-order atom $A$'s sum of weights to $n_A(S)$, the number of groundings of $A$ compatible with the constraints in $S$. This is necessary and sufficient since each atom $A$ has $n_A(S)$ groundings, and all ground atoms are independent because the CNF is satisfied irrespective of their truth values. Note that $n_A(S)$ is the number of groundings of $A$ consistent with $S$ that can be formed using all the constants in the original CNF.

## 5.2 LIFTING THE DECOMPOSITION STEP

Clearly, if $C$ can be decomposed into two or more CNFs such that no two CNFs share any unifiable literals, a lifted decomposition of $C$ is possible (i.e., a decomposition of $C$ into first-order CNFs on which LWMC can be called recursively). But the symmetry of the first-order representation can be further exploited. For example, if the CNF $C$ can be decomposed into $k$ CNFs $C_1, \ldots, C_k$ sharing no unifiable literals and such that for all $i, j$, $C_i$ is identical to $C_j$ up to a renaming of the variables and constants,[1] then LWMC$(C) = [\text{LWMC}(C_1)]^k$. We formalize these conditions below.

**Definition 1** *The set of first-order CNFs $\{C_{1,1}, \ldots, C_{1,m_1}, \ldots, C_{k,1}, \ldots, C_{k,m_k}\}$ is called a lifted decomposition of CNF $C$ under substitution constraints $S$ if, given $S$, it satisfies the following three properties: (i) $C = C_{1,1} \wedge \ldots \wedge C_{k,m_k}$; (ii) no two $C_{i,j}$'s share any unifiable literals; (iii) for all $i, j, j'$, such that $j \neq j'$, $C_{i,j}$ is identical to $C_{i,j'}$*

**Proposition 1** *If $\{C_{1,1}, \ldots, C_{1,m_1}, \ldots, C_{k,1}, \ldots, C_{k,m_k}\}$ is a lifted decomposition of $C$ under $S$, then*

$$\text{LWMC}(C, S, W) = \prod_{i=1}^{k} [\text{LWMC}(C_{i,1}, S, W)]^{m_i}. \quad (3)$$

Rules for identifying lifted decompositions can be derived in a straightforward manner from the inversion argument in de Salvo Braz [10] and the power rule in Jha et al. [20]. An example of such a rule is given in the definition and proposition below. Note that this rule is more general than de Salvo Braz's inversion elimination [10].

**Definition 2** *A set of variables $X = \{x_1, \ldots, x_m\}$ is called a decomposer of a CNF $C$ if it has the following three properties: (i) $X$ is the union of all variables appearing as the same argument of a predicate R in $C$; (ii) every $x_i$ in $X$ appears in all atoms of a clause in $C$; (iii) if $x_i$ and $x_j$ appear as arguments of a predicate R', they must appear as the same argument of R'. (R' may or may not be the same as R.)*

For example, $\{x_1, x_2\}$ is a decomposer of the CNF $(R(x_1) \vee S(x_1, x_3)) \wedge (R(x_2) \vee T(x_2, x_4))$. Given a decomposer $\{x_1, \ldots, x_m\}$ and a CNF $C$, it is easy to see that for every pair of substitutions of the form $S_X = \{x_1 = X, \ldots, x_m = X\}$ and $S_Y = \{x_1 = Y, \ldots, x_m = Y\}$, with $X \neq Y$, the CNFs $C_X$ and $C_Y$ obtained by applying $S_X$ and $S_Y$ to $C$ do not share any unifiable literals. A decomposer thus yields a lifted decomposition. Given a CNF, a decomposer can be found in linear time.

When there are no substitution constraints on the variables in the decomposer, as in the example above, all CNFs formed by substituting the variables in the decomposer with a constant are identical. Thus, $k = 1$ in Equation 3 and $m_1$ equals the number of constants (objects) in the PKB. However, when there are substitution constraints, the CNFs may not be identical. For example, given the CNF $(R(x_1) \vee S(x_1, x_3)) \wedge (R(x_2) \vee T(x_2, x_4))$ and substitution constraints $\{x_1 \neq A, x_2 \neq B\}$, the CNF formed by substituting $\{x_1 = A, x_2 = B\}$ is not identical to the CNF formed by substituting $\{x_1 = C, x_2 = C\}$.

---
[1]Henceforth, when we say that two clauses are identical, we mean that they are identical up to a renaming of constants and variables.

Intuitively, if all the clauses in the CNF have the same set of groundings relative to the decomposer, then any two CNFs formed by substituting the variables in the decomposer with any two (valid) distinct constants will be identical. Thus, we need to split the CNF into disjoint CNFs that have identical groundings relative to the decomposer. We can achieve this by considering all possible combinations of the substitution constraints. For instance, we can decompose our example CNF into the following four CNFs, each of which has an identical set of groundings relative to $x_1$ and $x_2$ (for readability, we do not standardize variables apart and show the constraints separately for each CNF): (1) $(R(x_1) \vee S(x_1, x_3)) \wedge (R(x_2) \vee T(x_2, x_4))$, $\{x_1 \neq A, x_1 \neq B, x_2 \neq A, x_2 \neq B\}$; (2) $(R(x_1) \vee S(x_1, x_3)) \wedge (R(x_2) \vee T(x_2, x_4))$, $\{x_1 \neq A, x_1 = B, x_2 \neq A, x_2 = B\}$; (3) $(R(x_1) \vee S(x_1, x_3)) \wedge (R(x_2) \vee T(x_2, x_4))$, $\{x_1 = A, x_2 = A, x_1 \neq B, x_2 \neq B\}$; and (4) $(R(x_1) \vee S(x_1, x_3)) \wedge (R(x_2) \vee T(x_2, x_4))$, $\{x_1 = A, x_1 = B, x_2 = A, x_2 = B\}$. Notice that the fourth CNF has no valid groundings and can be removed.

In general, a CNF can be partitioned into subsets of identical but disjoint CNFs using constraint satisfaction techniques, as in Kisynski and Poole [22]. In summary:

**Proposition 2** *Let $X = \{x_1, \ldots, x_t\}$ be a decomposer of $C$. Let $\{\{X_{1,1}, \ldots, X_{1,m_1}\}, \ldots, \{X_{k,1}, \ldots, X_{k,m_k}\}\}$ be a partition of the constants in the domain and let $C' = \{C_{X_{1,1}}, \ldots, C_{X_{1,m_1}}, \ldots, C_{X_{k,1}}, \ldots, C_{X_{k,m_k}}\}$ be a partition of $C$ such that (i) for all $i, j, j'$ such that $j \neq j'$, $C_{X_{i,j}}$ is identical to $C_{X_{i,j'}}$, and (ii) $C_{X_{i,m_i}}$ is a CNF formed by substituting each variable in $X$ by the constant $X_{i,m_i}$. Then $C'$ is a lifted decomposition of $C$ under $S$.*

## 5.3 LIFTING THE SPLITTING STEP

Splitting on a non-ground atom means splitting on all groundings of it consistent with the current substitution constraints $S$. Naively, if the atom has $c$ groundings consistent with $S$ this will lead to a sum of $2^c$ recursive calls to LWMC, one for each possible truth assignment to the $c$ ground atoms. However, in general these calls will have repeated structure and can be replaced by a much smaller number. The lifted splitting step exploits this.

We introduce some notation and definitions. Let $\sigma_{A,S}$ denote a truth assignment to the groundings of atom $A$ consistent with substitution constraints $S$, and let $\Sigma_{A,S}$ denote the set of all possible such assignments. Let $C|\sigma_{A,S}$ denote the CNF formed by removing $\neg A$ from all clauses it appears in, and setting to *Satisfied* all ground clauses that are satisfied because of $\sigma_{A,S}$. This can be done in a lifted manner by updating the substitution constraints asso-ciated with each clause. For instance, consider the clause R(x) ∨ S(x, y) and substitution constraint {x ≠ A}, and suppose the domain is {A, B, C} (i.e., these are all the constants appearing in the PKB). Given the assignment R(A) = False, R(B) = True, R(C) = False and ignoring satisfied clauses, the clause becomes S(x, y) and the constraint set becomes {x ≠ A, x ≠ B}. R(x) is removed from the clause because all of its groundings are instantiated. The constraint x ≠ B is added because the assignment R(B) = True satisfies all groundings in which x = B.

**Definition 3** *The partition* $\{\Sigma_{A,S}^{(1)}, \ldots, \Sigma_{A,S}^{(l)}\}$ *of* $\Sigma_{A,S}$ *is called a* lifted split *of atom $A$ for CNF $C$ under substitution constraints $S$ if every part $\Sigma_{A,S}^{(i)}$ satisfies the following two properties: (i) all truth assignments in $\Sigma_{A,S}^{(i)}$ have the same number of true atoms; (ii) for all pairs $\sigma_j, \sigma_k \in \Sigma_{A,S}^{(i)}$, $C|\sigma_j$ is identical to $C|\sigma_k$.*

**Proposition 3** *If $\{\Sigma_{A,S}^{(1)}, \ldots, \Sigma_{A,S}^{(l)}\}$ is a lifted split of $A$ for $C$ under $S$, then*

$$\text{LWMC}(C, S, W) = \sum_{i=1}^{l} n_i W_A^{t_i} W_{\neg A}^{f_i} \text{LWMC}(C|\sigma_j; S_j, W)$$

*where $n_i = |\Sigma_{A,S}^{(i)}|$, $\sigma_j \in \Sigma_{A,S}^{(i)}$, $t_i$ and $f_i$ are the number of true and false atoms in $\sigma_j$, respectively, and $S_j$ is $S$ augmented with the substitution constraints required to form $C|\sigma_j$.*

Again, we can derive rules for identifying a lifted split by using the counting arguments in de Salvo Braz [10] and the generalized binomial rule in Jha et al. [20]. We omit the details for lack of space. In the worst case, lifted splitting defaults to splitting on a ground atom. In most inference problems, the PKB will contain many hard ground unit clauses (the evidence). Splitting on the corresponding ground atoms then reduces to a single recursive call to LWMC for each atom. In general, the atom to split on in Algorithm 5 should be chosen with the goal of yielding lifted decompositions in the recursive calls (for example, using lifted versions of the propositional heuristics [35]).

Notice that the lifting schemes used for decomposition and splitting in Algorithm 5 by no means exhaust the space of possible probabilistic lifting rules. For example, Jha et al. [20] and Milch et al. [24] contain examples of other lifting rules. Searching for new probabilistic lifted inference rules, and positive and negative theoretical results about what can be lifted, looks like a fertile area for future research.

The theorem below follows from Theorem 2 and the arguments above.

**Theorem 3** *Algorithm LWMC($C$, ∅, $W$) correctly computes the weighted model count of CNF $C$ under literal weights $W$.*

### 5.4 EXTENSIONS

Although most probabilistic logical languages do not include existential quantification, handling it in PTP is desirable for the sake of logical completeness. This is complicated by the fact that skolemization is not sound for model counting (skolemization will not change satisfiability but can change the model count), and so cannot be applied. The result of conversion to CNF is now a conjunction of clauses with universally and/or existentially quantified variables (e.g., [∀x∃y (R(x, y) ∨ S(y))] ∧ [∃u∀v∀w T(u, v, w)]). Algorithm 5 now needs to be able to handle clauses of this form. If no universal quantifier appears nested inside an existential one, this is straightforward, since in this case an existentially quantified clause is just a compact representation of a longer one. For example, if the domain is {A, B, C}, the unit clause ∀x∃y R(x, y) represents the clause ∀x (R(x, A) ∨ R(x, B) ∨ R(x, C)). The decomposition and splitting steps in Algorithm 5 are both easily extended to handle such clauses without loss of lifting (and the base case does not change). However, if universals appear inside existentials, a first-order clause now corresponds to a disjunction of conjunctions of propositional clauses. For example, if the domain is {A, B}, ∃x∀y (R(x, y) ∨ S(x, y)) represents (R(A, A)∨S(A, A))∧(R(A, B)∨S(A, B))∨(R(B, A)∨ S(B, A)) ∧ (R(B, B) ∨ S(B, B)). Whether these cases can be handled without loss of lifting remains an open question.

Several optimizations of the basic LWMC procedure in Algorithm 5 can be readily ported from the algorithms PTP generalizes. These optimizations can tremendously improve the performance of LWMC.

**Unit Propagation** When LWMC splits on atom $A$, the clauses in the current CNF are resolved with the unit clauses $A$ and $\neg A$. This results in deleting false atoms, which may produce new unit clauses. The idea in unit propagation is to in turn resolve all clauses in the new CNF with all the new unit clauses, and continue to do this until no further unit resolutions are possible. This often produces a much smaller CNF, and is a key component of DPLL that can also be used in LWMC. Other techniques used in propositional inference that can be ported to LWMC include pure literals, clause learning, clause indexing, and random restarts [3, 35, 4].

**Caching/Memoization** In a typical inference, LWMC will be called many times on the same subproblems. The solutions of these can be cached when they are computed, and reused when they are encountered again. (Notice that a cache hit only requires identity up to renaming of variables and constants.) This can greatly reduce the time complexity of LWMC, but at the cost of increased space complexity. If the results of all calls to LWMC are cached (full caching), in the worst case LWMC will use exponential space. In practice, we can limit the cache size to the available memory and heuristically prune elements from it when it is full. Thus, by changing the cache size, LWMC can explore various time/space tradeoffs. Caching in LWMC corresponds to both caching in model counting [35] and recursive conditioning [5] and to memoization of common subproofs in theorem proving [39].

**Knowledge-Based Model Construction** KBMC first uses logical inference to select the subset of the PKB that is rel-evant to the query, and then propositionalizes the result and performs standard probabilistic inference on it [40]. A similar effect can be obtained in PTP by noticing that in Equation 2 factors that are common to $Z(K \cup \{(Q, 0)\})$ and $Z(K)$ cancel out and do not need to be computed. Thus we can modify Algorithm 3 as follows: (i) simplify the PKB by unit propagation starting from evidence atoms, etc.; (ii) drop from the PKB all formulas that have no path of unifiable literals to the query; (iii) pass to LWMC only the remaining formulas, with an initial $S$ containing the substitutions required for the unifications along the connecting path(s).

### 5.5 THEORETICAL PROPERTIES

We now theoretically compare the efficiency of PTP and first-order variable elimination (FOVE) [29, 10].

**Theorem 4** *PTP can be exponentially more efficient than FOVE.*

*Proof sketch.* We provide a constructive proof. Consider the hard CNF $(R_1(x_1) \vee R_2(x_1, x_2) \vee R_3(x_2, x_3)) \wedge (\neg R_1(x_1) \vee R_2(x_2, x_1) \vee R_4(x_2, x_3)) \wedge (R_1(x_1))$. Neither counting elimination [10] nor inversion elimination [29] is applicable here, and therefore the complexity of FOVE will be the same as that of (propositional) variable elimination, i.e., exponential in the treewidth. The treewidth of the CNF is polynomial in the domain size (number of constants), and therefore variable elimination and by extension FOVE will require exponential time. On the other hand, PTP will solve this problem in polynomial time. Since $R_1(x_1)$ is a unit clause, PTP will remove the first clause because it is satisfied (clause deletion). It will then remove $R_1$ from the second clause (unit propagation), yielding the hard CNF $R_2(x_2, x_1) \vee R_4(x_2, x_3)$. PTP will solve this reduced CNF by first running lifted decomposition ($x_2$ is a decomposer) followed by two lifted splits over $R_2(A, x_1)$ and $R_3(A, x_3)$. Thus, the overall time complexity of PTP is polynomial in the domain size. □

**Theorem 5** *LWMC with full caching has the same worst-case time and space complexity as FOVE.*

The proof of Theorem 5 is a little involved. The main insight for this result comes from previous work on recursive conditioning [5] and AND/OR search [12]. Specifically, these papers show that the worst-case time and space complexity of propositional WMC with caching (and without unit propagation and clause deletion) is the same as that of variable elimination (VE) (exponential in the treewidth). Specifically the authors show that both WMC and VE are graph traversal schemes that operate by traversing the same graph in a top-down and bottom-up manner respectively. Lifting can be seen as a way of compressing this graph via Propositions 1 and 3; specifically by aggregating nodes in the graph that behave similarly. Since FOVE is a lifted analog of VE, it traverses the compressed lifted graph in a bottom-up manner while LWMC with caching traverses it in a top-down manner; assuming that they use the same rules for lifting. Since the lifting rules used by LWMC are at least as general as FOVE, its worst-case time and space complexity is the same as FOVE.

De Salvo Braz's FOVE [10] and lifted BP [38] completely shatter the PKB in advance. This may be wasteful because many of those splits may not be necessary. Like Poole [29] and Ng et al. [26], LWMC splits only as needed.

### 5.6 DISCUSSION

PTP yields new algorithms for several of the inference problems in Figure 1. For example, ignoring weights and replacing products by conjunctions and sums by disjunctions in Algorithm 5 yields a lifted version of DPLL for first-order theorem proving (cf. [2]).

Of the standard methods for inference in graphical models, propositional PTP is most similar to recursive conditioning [5] and AND/OR search [12] with context-sensitive decomposition and caching, but applies to arbitrary PKBs, not just Bayesian networks. Also, PTP effectively performs formula-based inference [17] when it splits on one of the auxiliary atoms introduced by Algorithm 2.

PTP realizes some of the benefits of lazy inference for relational models [31] by keeping in lifted form what lazy inference would leave as default.

## 6 APPROXIMATE INFERENCE

LWMC lends itself readily to Monte Carlo approximation, by replacing the sum in the splitting step with a random choice of one of its terms, calling the algorithm many times, and averaging the results. This yields the first lifted sampling algorithm.

We first apply this importance sampling approach [33] to WMC, yielding the MC-WMC algorithm. The two algorithms differ only in the last line. Let $Q(A|C, W)$ denote the *importance* or *proposal* distribution over $A$ given the current CNF $C$ and literal weights $W$. Then we return $\frac{W_A}{Q(A|C,W)}$MC-WMC$(C|A; W)$ with probability $Q(A|C, W)$, or $\frac{W_{\neg A}}{Q(\neg A|C,W)}$MC-WMC$(C|\neg A; W)$ otherwise. By importance sampling theory [33] and by the law of total expectation, it is easy to show that:

**Theorem 6** *If $Q(A|C, W)$ satisfies* WMC$(C|A; W) > 0$ $\Rightarrow Q(A|C, W) > 0$ *for all atoms $A$ and its true and false assignments, then the expected value of the quantity output by* MC-WMC$(C, W)$ *equals* WMC$(C, W)$. *In other words,* MC-WMC$(C, W)$ *yields an unbiased estimate of* WMC$(C, W)$.

An estimate of WMC$(C, W)$ is obtained by running MC-WMC$(C, W)$ multiple times and averaging the results. By linearity of expectation, the running average is also unbiased. It is well known that the accuracy of the estimate is inversely proportional to its variance [33]. The variance can be reduced by either running MC-WMC more times or by choosing $Q$ that is as close as possible to the posterior distribution $P$ (or both). Thus, for MC-WMC to be effective in practice, at each point, given the current CNF $C$, we should select $Q(A|C, W)$ that is as close as possible to the marginal probability distribution of $A$ w.r.t. $C$ and $W$.

The following simple procedure can be used to construct the proposal distribution $Q$. Let $A$ be an atom that needs to be sampled and (abusing notation) let $o = (A_1, \ldots, A_n)$ be an ordering of its ground atoms (we select the ordering randomly). Given a truth assignment to the previous $i - 1$ atoms, let $n_{i,t}$ and $n_{i,f}$ denote the number of ground clauses that are satisfied by assigning $A_i$ to true and false respectively. Then, we use $Q(A_i|A_1, \ldots, A_{i-1}, C, W) = n_{i,t} W_A / (n_{i,t} W_A + n_{i,f} W_{\neg A})$. Thus we perform only a one-step look ahead for constructing $Q$. In future, we envision using more sophisticated heuristics.

MC-WMC suffers from the rejection problem [18]: it may return a zero. We can solve this problem by either backtracking when a sample is rejected or by generating samples from the backtrack-free distribution [18].

Next, we present a lifted version of MC-WMC, which is obtained by replacing the (last line of the) lifted splitting step in LWMC by the following lifted sampling step:

**return** $\frac{n_i W_A^{t_i} W_{\neg A}^{f_i}}{Q(\Sigma_{A,S}^{(i)})}$MC-LWMC$(C|\sigma_j; S_j, W)$
 where $n_i$, $t_i$, $f_i$, $\sigma_j$ and $S_j$ are as in Proposition 3

In the lifted sampling step, we construct a distribution $Q$ over the lifted split and sample an element $\Sigma_{A,S}^{(i)}$ from it. Then we weigh the sampled element w.r.t. $Q$ and call the algorithm recursively on the CNF conditioned on $\sigma_j \in \Sigma_{A,S}^{(i)}$. Notice that $A$ is a first-order atom and the distribution $Q(\Sigma_{A,S}^{(i)})$ is defined in a lifted manner. However, semantically, each $\Sigma_{A,S}^{(i)}$ represents all of groundings of $A$ and therefore given a ground assignment $\sigma_j \in \Sigma_{A,S}^{(i)}$, the probability of sampling $\sigma_j$ is $Q_G(\sigma_j) = Q(\Sigma_{A,S}^{(i)})/n_i$. Thus, ignoring the decomposition step, MC-LWMC is equivalent to MC-WMC that uses $Q_G$ to sample all the groundings of $A$. In the decomposition step, given a set of identical and disjoint CNFs, we simply sample just one of the CNFs and raise our estimate to the appropriate count. The correctness of this step follows from the fact that the expected value of the product of $k$ identical and independent random variables $R_1, \ldots, R_k$ equals $E[R_1]^k$, and $(\sigma_{R_1})^k$ is an unbiased estimate of $E[R_1]^k$ where $\sigma_{R_1}$ is a random sample of $R_1$. Therefore, the following theorem immediately follows from Theorem 6.

**Theorem 7** *If $Q(\Sigma_{A,S}^{(i)})$ satisfies WMC$(C|\sigma_j; S_j, W) > 0 \Rightarrow Q(\Sigma_{A,S}^{(i)}) > 0$ for all elements $\Sigma_{A,S}^{(i)}$ of the lifted split of $A$ for $C$ under $S$, then MC-LWMC$(C, S, W)$ yields an unbiased estimate of WMC$(C, W)$.*

MC-LWMC has smaller variance than MC-WMC and is therefore likely to have higher accuracy. The smaller variance is due to the smaller time complexity of MC-LWMC, which in turn is due to the decomposition step. Recall that we group identical and independent CNFs, sample just one CNF from the group, and raise the estimate by the number of members in the group. Thus, for each lifted decomposition of size $m_i > 1$, we have a factor of $m_i$ speedup. Therefore, given a specific time bound, the estimate returned by MC-LWMC will be based on a larger sample size (or more runs) than the one returned by MC-WMC.

| #Objs. | 10 | | 20 | | 50 | |
|---|---|---|---|---|---|---|
| Clause Size ↓ | PTP | FOVE | PTP | FOVE | PTP | FOVE |
| 3 | 14.5 | 18.93 | 34.5 | 93.45 | 82.1 | X |
| 5 | 23.9 | X | 43.5 | X | 132.4 | X |
| 7 | 8.2 | X | 18.7 | X | 37.1 | X |
| 9 | 2.3 | X | 5.2 | X | 15.9 | X |

Table 1: Impact of increasing the number of objects and clause size on the time complexity of FOVE and PTP. Time is in seconds. 'X' indicates that the algorithm ran out of memory.

## 7 EXPERIMENTS

### 7.1 EXACT INFERENCE

In this subsection, we compare the performance of PTP and FOVE on randomly generated and link prediction PKBs. We implemented PTP in C++ and ran all our experiments on a Linux machine with a 2.33 GHz Intel Xeon processor and 2GB of RAM. We used a constraint solver based on forward checking to implement the substitution constraints. We used the following heuristics for splitting. At any point, we prefer an atom which yields the smallest number of recursive calls to LWMC (i.e., an atom that yields maximum lifting). We break ties by selecting an atom that appears in the largest number of ground clauses; this number can be computed using the constraint solver. If it is the same for two or more atoms, we break ties randomly.

**Random PKBs with Varying Clause Size** In the first set of experiments, we show that PTP's advantage relative to FOVE increases with clause length. In order to compare the performance in a controlled setting, we generated random PKBs parameterized by five integer constants: $n$, $m$, $s$, $e$ and $c$, where $n$ is the number of predicates, $m$ is the number of clauses, $s$ is the number of literals in each clause, $e$ is the number of evidence atoms, and $c$ is the number of constants in the domain. The PKB is generated as follows. All predicates are unary. We generate $m$ clauses by randomly selecting $s$ predicates and negating each with probability 0.5. We then choose $e$ ground atoms as evidence, each of which is set to either True or False with equal probability.

We set $n = m = 40$, varied $s$ from 3 to 9 in increments of 2 and $c$ from 10 to 50, and set $e = c/10$. Table 1 shows the impact of increasing the number of objects $c$ and the clause size $s$ on the time complexity of FOVE and PTP. The results are averaged over 10 PKBs. We can see that PTP always dominates FOVE. When the PKB has small clauses, PTP is only slightly better than FOVE. However, when the clauses are large, PTP is substantially better than FOVE, which runs out of memory on all the instances, typically after around 20 minutes of run time. When large clauses are present, unit propagation is very effective and causes a large amount of pruning. Because of this, PTP is much faster than FOVE.

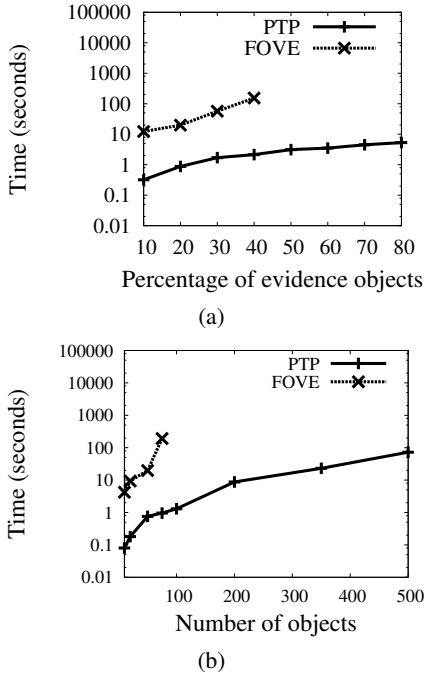

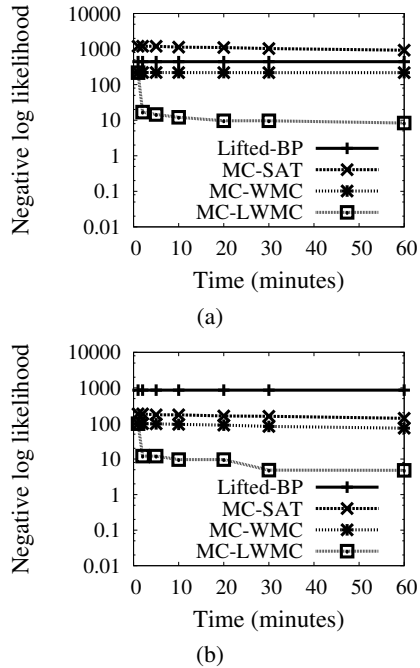

Figure 2: (a) Impact of increasing the amount of evidence on the time complexity of FOVE and PTP in the link prediction domain. The number of objects in the domain is 100. (b) Impact of increasing the number of objects on the time complexity of FOVE and PTP in the link prediction domain, with 20% of the atoms set as evidence.

Figure 3: Negative log-likelihood of the data as a function of time for lifted BP, MC-SAT, MC-WMC and MC-LWMC on (a) the entity resolution (Cora) and (b) the collective classification domains.

**Link Prediction** We experimented with a simple PKB consisting of two clauses: $\text{GoodProf}(x) \wedge \text{GoodStudent}(y) \wedge \text{Advises}(x,y) \Rightarrow \text{FutureProf}(y)$ and $\text{Coauthor}(x,y) \Rightarrow \text{Advises}(x,y)$. The PKB has two types of objects: professors ($x$) and students ($y$). Given data on a subset of papers and "goodness" of professors and students, the task is to be predict who advises whom and who is likely to be a professor in the future.

We evaluated the performance of FOVE and PTP along two dimensions: (i) the number of objects and (ii) the amount of evidence. We varied the number of objects from 10 to 1000 and the number of evidence atoms from 10% to 80%.

Figure 2(a) shows the impact of increasing the number of evidence atoms on the performance of the two algorithms on a link prediction PKB with 100 objects. FOVE runs out of memory (typically after around 20 minutes of run time) after the percentage of evidence atoms rises above 40%. PTP solves all the problems and is also much faster than FOVE (notice the log-scale on the y-axis). Figure 2(b) shows the impact of increasing the number of objects on a link prediction PKB with 20% of the atoms set as observed. We can see that FOVE is unable to solve any problems after the number of objects is increased beyond 100 because it runs out of memory. PTP, on the other hand, solves all problems in less than 100s.

### 7.2 APPROXIMATE INFERENCE

In this subsection, we compare the performance of MC-LWMC, MC-WMC, lifted belief propagation [38], and MC-SAT [30] on two domains. We used the entity resolution (Cora) and collective classification datasets and Markov logic networks used in Singla and Domingos [37] and Poon and Domingos [30] respectively. The Cora dataset contains 1295 citations to 132 different research papers. The inference task here is to detect duplicate citations, authors, titles and venues. The collective classification dataset consists of about 3000 query atoms.

Since computing the exact posterior marginals is infeasible in these domains, we used the following evaluation method. We partitioned the data into two equal-sized sets: evidence set and test set. We then computed the probability of each ground atom in the test set given all atoms in the evidence set using the four inference algorithms. We measure the error using negative log-likelihood of the data according to the inference algorithms (the negative log-likelihood is a sampling approximation of the K-L divergence to the data-generating distribution, shifted by its entropy).

The results, averaged over 10 runs, are shown in Figures 3(a) and 3(b). The figures show how the log-likelihood of the data varies with time for the four inference algorithms used. We see that MC-LWMC has the lowest negative log-likelihood of all algorithms by a large margin. It significantly dominates MC-WMC in about 2 minutes of run-time and is substantially superior to both lifted BP and MC-SAT

(notice the log scale). This shows the advantages of approximate PTP over lifted BP and ground inference.

## 8 CONCLUSION

Probabilistic theorem proving (PTP) combines theorem proving and probabilistic inference. This paper proposed an algorithm for PTP based on reducing it to lifted weighted model counting, and showed both theoretically and empirically that it has significant advantages compared to previous lifted probabilistic inference algorithms. An implementation of PTP will be available in the Alchemy system [25].

Directions for future research include: extension of PTP to infinite, non-Herbrand first-order logic; new lifted inference rules; theoretical analysis of liftability; porting to PTP more speedup techniques from logical and probabilistic inference; lifted splitting heuristics; better handling of existentials; variational PTP algorithms; better importance distributions; approximate lifting; answering multiple queries simultaneously; applications; etc.

**Acknowledgements** This research was partly funded by ARO grant W911NF-08-1-0242, AFRL contract FA8750-09-C-0181, DARPA contracts FA8750-05-2-0283, FA8750-07-D-0185, HR0011-06-C-0025, HR0011-07-C-0060 and NBCH-D030010, NSF grants IIS-0534881 and IIS-0803481, and ONR grant N00014-08-1-0670. The views and conclusions contained in this document are those of the authors and should not be interpreted as necessarily representing the official policies, either expressed or implied, of ARO, DARPA, NSF, ONR, or the U.S. Government.